\documentclass{article}

\usepackage{arxiv}
\usepackage{float}
\usepackage{amsmath}
\usepackage{amssymb} 
\usepackage[utf8]{inputenc} 
\usepackage[T1]{fontenc}    
\usepackage{hyperref}       
\usepackage{url}            
\usepackage{booktabs}       
\usepackage{amsfonts}       
\usepackage{nicefrac}       
\usepackage{microtype}      
\usepackage{graphicx}
\usepackage[numbers,sort&compress]{natbib}
\usepackage{doi}
\usepackage{authblk}
\usepackage{placeins}

\AtBeginDocument{%
  \setlength{\headheight}{23pt}%
  \addtolength{\topmargin}{-9pt}%
}

\renewcommand{\headeright}{}
\renewcommand{\undertitle}{}

\date{} 
\title{A Comparative Analysis of Machine Learning and Deep Learning Models for Tweet Sentiment Classification: A Case Study on the Sentiment 140 Dataset}
\renewcommand{\shorttitle}{Comparative Analysis of ML and DL for Tweet Sentiment Classification}

\author{
  \begin{minipage}[t]{0.45\textwidth}
    \centering
    \textbf{Vita Anggraini}\\[-0.2em]
    Faculty of Science\\[-0.2em]
    Sumatra Institute of Technology\\[-0.2em]
    \texttt{vita.122450046@student.itera.ac.id}
  \end{minipage}
  \hfill
  \begin{minipage}[t]{0.45\textwidth}
    \centering
    \textbf{Cintya Bella}\\[-0.2em]
    Faculty of Science\\[-0.2em]
    Sumatra Institute of Technology\\[-0.2em]
    \texttt{cintya.122450066@student.itera.ac.id}
  \end{minipage}

  \vspace{0.25cm}

  \begin{minipage}[t]{0.45\textwidth}
    \centering
    \textbf{Bastian}\\[-0.2em]
    Faculty of Science\\[-0.2em]
    Sumatra Institute of Technology\\[-0.2em]
    \texttt{bastian.122450130@student.itera.ac.id}
  \end{minipage}
  \hfill
  \begin{minipage}[t]{0.45\textwidth}
    \centering
    \textbf{Luluk Muthoharoh}\\[-0.2em]
    Faculty of Science\\[-0.2em]
    Sumatra Institute of Technology\\[-0.2em]
    \texttt{luluk.muthoharoh@sd.itera.ac.id}
  \end{minipage}

  \vspace{0.25cm}

  \begin{minipage}[t]{0.45\textwidth}
    \centering
    \textbf{Ardika Satria}\\[-0.2em]
    Faculty of Science\\[-0.2em]
    Sumatra Institute of Technology\\[-0.2em]
    \texttt{ardika.satria@sd.itera.ac.id}
  \end{minipage}
  \hfill
  \begin{minipage}[t]{0.45\textwidth}
    \centering
    \textbf{Martin C.T. Manullang}\\[-0.2em]
    Faculty of Science\\[-0.2em]
    Sumatra Institute of Technology\\[-0.2em]
    \texttt{martin.manullang@if.itera.ac.id}
  \end{minipage}
}

\hypersetup{
pdftitle={A Comparative Analysis of Machine Learning and Deep Learning Models for Tweet Sentiment Classification},
pdfsubject={Sentiment analysis on Sentiment140 dataset},
pdfauthor={Vita Anggraini, Cintya Bella, Bastian, Luluk Muthoharoh, Ardika Satria, Martin C.T. Manullang},
pdfkeywords={sentiment analysis, logistic regression, BiLSTM, Sentiment140, Streamlit},
}

\begin{document}
\maketitle

\begin{abstract}
The exponential growth of social media has created an urgent need for automated systems to analyze unstructured public sentiment in real-time. This study compares a traditional Logistic Regression model (TF-IDF) against a deep learning Bidirectional Long Short-Term Memory (BiLSTM) architecture using a 10,000-tweet subset of the Sentiment140 dataset. Experimental results demonstrate that Logistic Regression outperformed BiLSTM, achieving an accuracy of 73.5\% compared to 69.17\%, with the deep learning model showing signs of mild overfitting. The findings suggest that for medium-scale informal text data, classical machine learning with robust feature extraction can yield superior results over complex deep learning architectures. Finally, the optimal models were integrated into an interactive web application via Streamlit and hosted on Hugging Face Spaces for public accessibility.
\end{abstract}


\section{Introduction}
In today’s highly interconnected digital era, social media platforms such as Twitter (now officially known as X) have irrevocably transformed the global communication landscape. These platforms serve as the primary medium for the global community to express their thoughts, voice complaints, and share emotions on a wide variety of topics. These topics range from critical government policies and political elections to product reviews, brand perception, and entertainment trends. Twitter, with its microblogging nature and character limits, forces users to express themselves concisely, often leading to the use of slang, abbreviations, emoticons, and implicit sarcasm.

This massive, unstructured, and continuous surge of opinionated text has triggered a profound need for Natural Language Processing (NLP) techniques. Manual analysis of such vast amounts of data is computationally impossible, economically unviable, and highly prone to human bias. Consequently, automated text processing, particularly sentiment analysis (also known as opinion mining), has become essential\cite{Rajput2020}. Sentiment analysis aims to computationally classify textual messages into specific affective polarities---most commonly binary categories such as positive and negative, or sometimes ternary categories including neutral sentiment.

Historically, initial approaches to solving the sentiment classification problem relied heavily on lexicon or dictionary-based techniques. These early systems matched words in a text against a pre-compiled dictionary of polarized words to determine the overall sentiment score. However, lexicon-based approaches struggled significantly with the dynamic nature of internet slang, context-dependent meanings, and linguistic nuances such as double negatives or irony. Over time, approaches based on artificial intelligence algorithms---encompassing both conventional machine learning and modern deep learning paradigms---have emerged as the undisputed industry standard. These data-driven algorithms learn complex patterns directly from annotated datasets, granting them the flexibility to map informal human language into actionable computational representations \cite{Muttaqin2021}. 

Despite the recent dominance of deep learning models in various NLP tasks, there remains an ongoing debate regarding their efficacy on smaller datasets compared to traditional, statistically grounded machine learning algorithms. Deep learning models require massive amounts of data to optimize their millions of parameters without falling into the trap of overfitting. Conversely, traditional machine learning models, when paired with robust feature extraction techniques, often demonstrate remarkable resilience and efficiency.

Therefore, this study aims to conduct an in-depth sentiment analysis on a dataset of tweets by designing, building, and rigorously comparing two distinct classification paradigms: a classical Logistic Regression model and a Bidirectional Long Short-Term Memory (BiLSTM) neural network model. Through this research, we aim to address the following objectives:
\begin{itemize}
    \item To implement a complete NLP pipeline, from raw text preprocessing to feature extraction and model training.
    \item To evaluate and compare the effectiveness, precision, and limitations of Logistic Regression (with TF-IDF) against a BiLSTM architecture when dealing with a constrained subset of informal text data.
    \item To analyze the training dynamics of the neural network to identify potential bottlenecks such as overfitting or vanishing gradients.
    \item To deploy the trained models into a production-ready interactive web interface, bridging the gap between theoretical data science and practical software engineering.
\end{itemize}

\section{Related Work}

\subsection{Sentiment Analysis in Microblogging}
Sentiment analysis is a fundamental sub-field of computational linguistics that functions to recognize, extract, and study subjective opinions, attitudes, and affective states from textual sources. In the context of microblogging platforms like Twitter, text data is typically characterized by short sentence structures, phonetic spellings, numerous acronyms, and rich emoticon usage. Researchers have long identified that standard NLP tools trained on formal texts (like news articles or Wikipedia) perform poorly on tweets. Computational systems are tasked with categorizing these utterances into polarities, which has proven crucial in domains such as business intelligence, stock market prediction, and monitoring public response during crises \cite{Nakov2016}. 

\subsection{Feature Engineering for Text Classification}
In traditional machine learning, algorithms cannot process raw text strings directly; the text must be converted into numerical vectors. Bag-of-Words (BoW) was an early method, but it suffered from giving equal weight to all words, regardless of their semantic importance. Term Frequency-Inverse Document Frequency (TF-IDF) emerged as a superior statistical feature extraction technique. It evaluates how important a word is to a document within a corpus, ensuring that words that appear rarely globally but frequently in a specific document receive a higher weight value \cite{Salton1988}. 

This approach creates highly sparse, high-dimensional vectors that are particularly well-suited for linear classifiers.

\subsection{Traditional Machine Learning Approaches}
Various traditional machine learning algorithms have been utilized for text classification, including Support Vector Machines (SVM), Naive Bayes, and Logistic Regression. Despite having the word "regression" in its nomenclature, Logistic Regression is a highly robust, mathematically sound algorithm for solving binary classification problems. Research by Hendriyana et al. (2022) highlights the efficiency of Logistic Regression in handling linearly separable data in high-dimensional spaces without the extreme computational overhead required by more complex algorithms \cite{Hendriyana2022}. 

\subsection{The Rise of Deep Learning and BiLSTM}
To overcome the limitations of sparse representations (like TF-IDF) which ignore word order and context, deep learning models utilize dense vector representations (word embeddings). Recurrent Neural Networks (RNNs) were initially used to process sequential data, but they suffered from the vanishing gradient problem. Long Short-Term Memory (LSTM) networks solved this by introducing gating mechanisms\cite{Hochreiter1997}. 

Bidirectional LSTM (BiLSTM) is an advanced innovation of the standard LSTM. A BiLSTM processes data from two directions simultaneously (forwards and backwards), allowing the network to capture both the past linguistic context and the future context, thereby providing a much more complete and accurate semantic understanding of a sentence before making a classification decision \cite{Alghifari2022}. 

\section{Theoretical Background}
To fully comprehend the experimental setup, it is necessary to detail the mathematical formulations governing the algorithms utilized in this study.

\subsection{Term Frequency-Inverse Document Frequency (TF-IDF)}
The TF-IDF weight is composed of two terms. The first computes the normalized Term Frequency (TF), which is the number of times a word $t$ appears in a document $d$, divided by the total number of words in that document.
\begin{equation}
    TF(t,d) = \frac{f_{t,d}}{\sum_{t' \in d} f_{t',d}}
\end{equation}

The second term is the Inverse Document Frequency (IDF), computed as the logarithm of the number of the documents in the corpus $N$ divided by the number of documents where the specific term appears.
\begin{equation}
    IDF(t,D) = \log \frac{N}{|\{d \in D: t \in d\}|}
\end{equation}

The final TF-IDF score is the product of these two metrics:
\begin{equation}
    TF-IDF(t,d,D) = TF(t,d) \times IDF(t,D)
\end{equation}

\subsection{Logistic Regression}
Logistic Regression predicts the probability $P$ that an observation belongs to a specific class (e.g., $y=1$ for positive sentiment) by fitting data to a logistic (sigmoid) function. The hypothesis formulation is:
\begin{equation}
    h_{\theta}(x) = \frac{1}{1 + e^{-(\theta^{T}x)}}
\end{equation}
Where $x$ represents the feature vector (TF-IDF scores) and $\theta$ represents the model weights. The model learns by minimizing the binary cross-entropy cost function:
\begin{equation}
    J(\theta) = -\frac{1}{m} \sum_{i=1}^{m} [y^{(i)} \log(h_{\theta}(x^{(i)})) + (1 - y^{(i)}) \log(1 - h_{\theta}(x^{(i)}))]
\end{equation}

\subsection{Bidirectional LSTM (BiLSTM) Architecture}
An LSTM cell contains three gates: forget gate ($f_t$), input gate ($i_t$), and output gate ($o_t$). These gates control the flow of information. For a given time step $t$, input $x_t$ and previous hidden state $h_{t-1}$, the cell state $C_t$ and hidden state $h_t$ are updated as follows:
\begin{align}
    f_t &= \sigma(W_f \cdot [h_{t-1}, x_t] + b_f) \\
    i_t &= \sigma(W_i \cdot [h_{t-1}, x_t] + b_i) \\
    \tilde{C}_t &= \tanh(W_c \cdot [h_{t-1}, x_t] + b_c) \\
    C_t &= f_t * C_{t-1} + i_t * \tilde{C}_t \\
    o_t &= \sigma(W_o \cdot [h_{t-1}, x_t] + b_o) \\
    h_t &= o_t * \tanh(C_t)
\end{align}
Where $\sigma$ is the sigmoid activation function and $W, b$ are the learnable weight matrices and biases. In a BiLSTM, the sequence is processed forward to obtain $\vec{h}_t$ and backward to obtain $\gets{h}_t$\cite{Schuster1997}. The final representation at time step $t$ is the concatenation of both hidden states:
\begin{equation}
    H_t = [\vec{h}_t \oplus \gets{h}_t]
\end{equation}

\section{Methodology}

\subsection{Dataset Acquisition and Sampling}
This study utilizes a secondary data source, namely the well-known public dataset "Sentiment140" obtained through the Kaggle platform \cite{Go2009}. The original dataset contains 1.6 million tweets automatically annotated using distant supervision (leveraging emoticons to determine initial labels).

To accommodate computational limits and accelerate the experimental iteration process, we employed a random sampling technique to extract a representative subset of exactly 10,000 observational data points. The dataset is strictly binary, mapped as shown in Table 1.

\begin{table}[H]
    \caption{Distribution of the Sampled Sentiment140 Dataset}
    \centering
    \begin{tabular}{lll}
        \toprule
        Class Label & Sentiment Polarity & Description \\
        \midrule
        Class 0 & Negative & Contains complaints, sadness, or anger \\
        Class 1 & Positive & Contains joy, appreciation, or support \\
        \bottomrule
    \end{tabular}
\end{table}

\subsection{Data Preprocessing Pipeline}
Raw text data from Twitter is highly noisy. We implemented a rigorous data preprocessing pipeline to standardize the input before it reaches the models\cite{Elmaliyasari2025}. The sequential steps are:
\begin{enumerate}
    \item \textbf{Case Folding}: All alphabetic characters are converted into lowercase to ensure that words like "Happy" and "happy" are treated as the identical semantic entity.
    \item \textbf{Data Cleaning}: We utilized Regular Expressions (Regex) to strip the text of hyperlinks ($http/https$), Twitter user mentions (@user), hashtags (\#), numerical digits, punctuation marks, and non-ASCII special characters.
    \item \textbf{Tokenization}: The continuous string of cleaned text is broken down into a list of discrete constituent tokens (words).
    \item \textbf{Padding and Truncation (For DL only)}: Neural networks require inputs of uniform tensor dimensions. We established a maximum sequence length limit of 50 tokens.
    \item \textbf{Vocabulary Creation}: A word index dictionary is compiled from all unique tokens found in the training dataset, resulting in a vocabulary size of 4,425 unique tokens.
\end{enumerate}

\subsection{Model Architecture and Hyperparameters}
The system executes two primary architectures. The baseline machine learning model utilizes the Scikit-Learn library\cite{Pedregosa2011}, employing a TF-IDF Vectorizer fitted on the training corpus, subsequently feeding a Logistic Regression classifier using the default L2 penalty.

The deep learning model is constructed using PyTorch\cite{Paszke2019}. The architectural topology and hyperparameter configurations are summarized in Table 2.

\begin{table}[H]
    \caption{BiLSTM Architecture and Hyperparameter Configuration}
    \centering
    \begin{tabular}{ll}
        \toprule
        Parameter / Component & Value / Specification \\
        \midrule
        Vocabulary Size & 4,425 \\
        Embedding Dimension & 128 \\
        Hidden Dimension ($hidden\_dim$) & 128 \\
        Number of LSTM Layers ($num\_layers$) & 2 \\
        Bidirectional & True \\
        Dropout Probability & 0.3 \\
        Max Sequence Length & 50 \\
        Total Trainable Parameters & 1,226,370 \\
        \midrule
        \textbf{Training Configuration} & \\
        Epochs & 6 \\
        Batch Size & 64 \\
        Optimizer & Adam \\
        Loss Function & Binary Cross-Entropy with Logits \\
        \bottomrule
    \end{tabular}
\end{table}

\subsection{Evaluation Metrics}
To evaluate the performance of the classification models quantitatively, we utilized a Confusion Matrix, which consists of four parameters: True Positive (TP), True Negative (TN), False Positive (FP), and False Negative (FN). Based on these parameters, we calculate standard evaluation metrics:

\begin{equation}
    Accuracy = \frac{TP + TN}{TP + TN + FP + FN}
\end{equation}

\begin{equation}
    Precision = \frac{TP}{TP + FP}
\end{equation}

\begin{equation}
    Recall = \frac{TP}{TP + FN}
\end{equation}

\begin{equation}
    F1-Score = 2 \times \frac{Precision \times Recall}{Precision + Recall}
\end{equation}

\section{Results and Discussion}

\subsection{Baseline Model Performance (Logistic Regression)}
Upon completion of the training phase, the Logistic Regression model was evaluated on the unseen testing set. Relying on the high-dimensional sparse representations generated by the TF-IDF feature extractor, the baseline model successfully achieved a robust accuracy rate of 73.5\%. This establishes a strong competitive benchmark, demonstrating the algorithm’s capability to draw effective linear decision boundaries in sparse semantic spaces.

\begin{figure}[H] 
    \centering
    \includegraphics[width=0.3\textwidth]{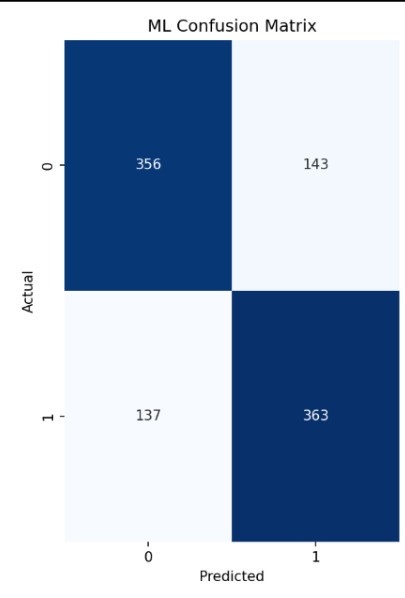} 
    \caption{Confusion Matrix for the Logistic Regression (Baseline) Model.}
    \label{fig:ml_cm}
\end{figure}

\FloatBarrier 

\subsection{Deep Learning Model Performance (BiLSTM)}
The evaluation results for the BiLSTM architecture yielded lower performance compared to the baseline. The detailed metrics are summarized in Table 3.

\begin{table}[H] 
    \centering
    \caption{Evaluation Metrics for the BiLSTM Model}
    \begin{tabular}{ll}
        \toprule
        Metric & Score (\%) \\
        \midrule
        Accuracy & 69.17\% \\
        Precision & 68.90\% \\
        Recall (Sensitivity) & 70.00\% \\
        F1-Score (Harmonic Mean) & 69.44\% \\
        \bottomrule
    \end{tabular}
    \label{tab:bilstm_metrics}
\end{table}

The exact distribution of predictions is further illustrated by the values from the Confusion Matrix recorded during testing:

\begin{figure}[H]
    \centering
    \includegraphics[width=0.4\textwidth]{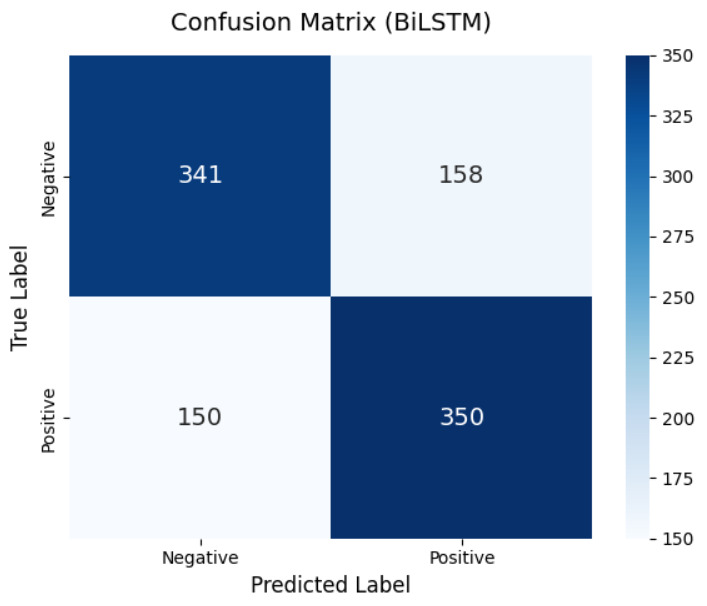} 
    \caption{Confusion Matrix for the BiLSTM Model.}
    \label{fig:dl_cm}
\end{figure}

\subsection{Comparative Analysis and Learning Curve Evaluation}
The empirical results reveal a significant phenomenon: the traditional Machine Learning baseline (Logistic Regression with TF-IDF) demonstrated superior stability and generalization compared to the computationally expensive BiLSTM model. 

To scientifically understand the behavior, generalization capability, and limitations of both architectures, we conducted a visual comparative analysis of their learning curves, as illustrated in Figure 3. The figure presents two distinct subplots: the left plot visualizes the Machine Learning learning curve as a function of training sample size, while the right plot displays the Deep Learning learning curve as a function of training epochs.

\textbf{Machine Learning Generalization (Left Plot):} Observing the Machine Learning curve, when the model is trained on a small subset of data (approximately 1,000 samples), it exhibits a natural high variance. The training accuracy starts exceptionally high at nearly 0.90, while the validation accuracy struggles at roughly 0.66. However, as the volume of training data increases up to 7,000 samples, a clear and healthy convergence occurs. The training accuracy gradually declines as the dataset becomes larger, while the validation accuracy rises steadily and eventually plateaus at approximately 74\%. The narrowing gap strongly signifies that adding more data successfully helped the Logistic Regression model generalize optimal linear decision boundaries without overfitting.

\textbf{Deep Learning Overfitting Phenomenon (Right Plot):} In stark contrast, the BiLSTM learning curve over 6 epochs vividly illustrates a critical bottleneck: overfitting due to data scarcity. The training accuracy demonstrates a steep upward trajectory, reaching nearly 0.95 by the final epoch, indicating that the network is effectively memorizing the training dataset. 

However, the validation accuracy fails to follow this upward trend, stagnating slightly above 0.70 after the second epoch. The progressively widening gap is the hallmark symptom of overfitting, where the BiLSTM network began to model specific noise and idiosyncrasies rather than extracting generalized linguistic rules.

This evaluation effectively proves that for medium-sized, highly informal text datasets, traditional Machine Learning models bolstered by TF-IDF reach a stable generalization point, while Deep Learning models require significantly larger datasets to avoid the memorization trap.
\begin{figure}[H]
    \centering
    \includegraphics[width=0.75\textwidth]{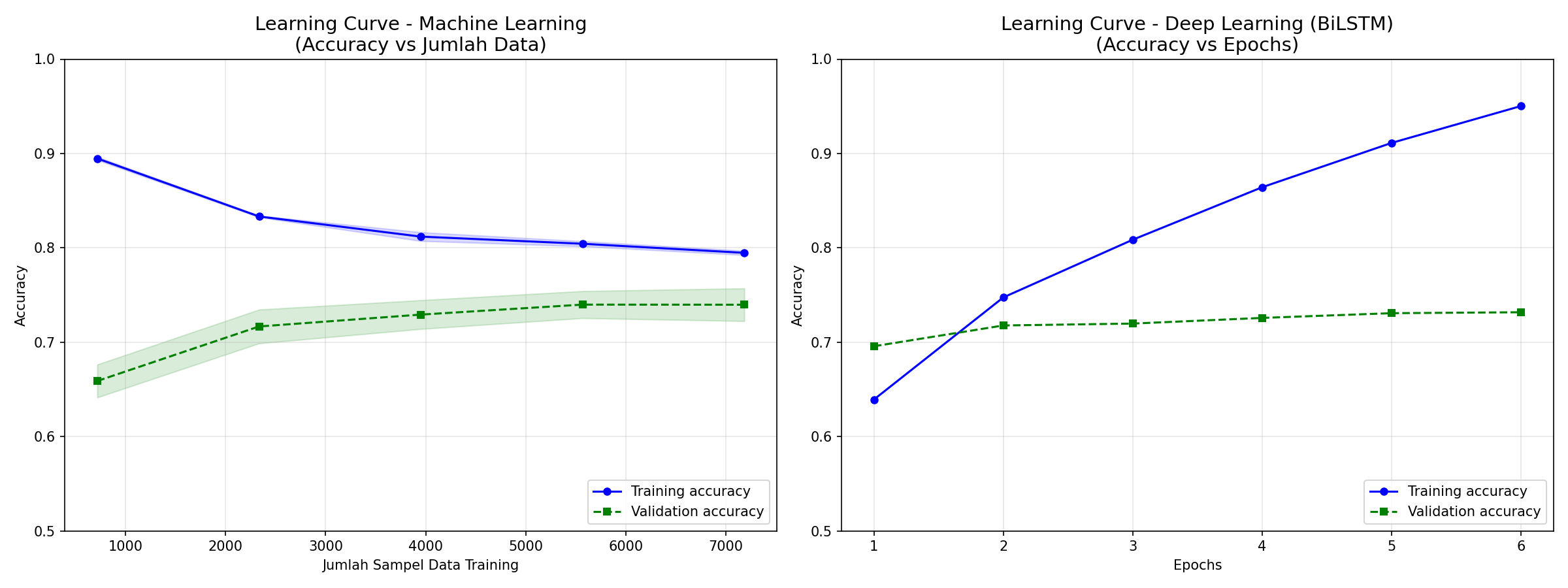} 
    \caption{Comparative Analysis of Learning Curves: (Left) Machine Learning Generalization, (Right) Deep Learning Overfitting Phenomenon.}
    \label{fig:learning_curves}
\end{figure}

\subsection{Model Deployment and Web Application}
A critical aspect of modern machine learning pipelines is the transition from a Jupyter Notebook environment to a production-ready application. To validate the real-world applicability of this study, both the Machine Learning (ML) and Deep Learning (DL) models were packaged alongside their necessary dependencies and deployed. These dependencies included the serialized model files: \texttt{bestsentimentmodel.pkl}, \texttt{tfidfvectorizer.pkl}, \texttt{bilstmstatedict.pt}, and \texttt{vocab.json}.

We utilized the Streamlit library to develop an intuitive graphical user interface (GUI). The source code was segregated into specific deployment folders (\texttt{huggingfaceapp/} for ML and \texttt{huggingfaceappdl/} for DL), allowing users to input arbitrary tweet text into a browser window and receive real-time sentiment predictions. This deployment effectively completes the end-to-end lifecycle of the project.

\section{Conclusion and Future Work}
This comprehensive study successfully designed, evaluated, and compared distinct computational architectures for informal text sentiment classification on Twitter data. Based on rigorous empirical testing, it is concluded that the classical Logistic Regression method combined with a TF-IDF feature extractor provides superior classification capabilities, achieving an accuracy of 73.5\%. In contrast, the BiLSTM neural network method only achieved a 69.17\% accuracy rate.

The decline in performance for the deep learning approach is highly attributable to the constrained volume of the training dataset (10,000 samples). The BiLSTM's massive parameter space led to mild overfitting, overshadowing its theoretical advantage in capturing sequential context. This reinforces the principle in data science that more complex algorithms do not universally guarantee better results, particularly when data is limited.

As a final product output, both approaches were successfully deployed into functional, interactive web applications within the Hugging Face Spaces ecosystem, proving their operational viability. For future research trajectories, several improvements are recommended. Firstly, scaling the training data to the full 1.6 million samples of the Sentiment140 dataset would likely allow the BiLSTM to surpass the linear baseline. Secondly, integrating pre-trained contextual embeddings such as Word2Vec, GloVe, or transitioning entirely to Transformer-based architectures like BERT (Bidirectional Encoder Representations from Transformers) could significantly enhance the semantic understanding of sarcastic or highly implicit tweets.






\bibliographystyle{unsrt} 
\bibliography{references}

\end{document}